\documentclass[conference]{IEEEtran}
\usepackage{times}

\usepackage[sort&compress,numbers]{natbib}
\usepackage{multicol}
\usepackage[
    bookmarks=true,
    hyperfootnotes=false
]{hyperref}

\usepackage{amsmath,amsfonts}
\usepackage[ruled,linesnumbered]{algorithm2e}

\usepackage{array}
\usepackage[caption=false,font=normalsize,labelfont=sf,textfont=sf]{subfig}
\usepackage{textcomp}
\usepackage{stfloats}
\usepackage{url}
\usepackage{verbatim}
\usepackage{graphicx}
\usepackage{xcolor}
\usepackage{siunitx}

\usepackage{multirow}

\usepackage[nolist,nohyperlinks]{acronym}
\renewcommand{\vec}[1]{\mathbf{#1}}

\begin{document}

\title{Efficient volumetric mapping of multi-scale environments using wavelet-based compression}

\author{\IEEEauthorblockN{Victor~Reijgwart, Cesar~Cadena, Roland~Siegwart and Lionel~Ott}
        \IEEEauthorblockA{Autonomous Systems Lab, ETH Z\"urich, Switzerland.}}

\maketitle

\begin{acronym}
\acro{SLAM}{Simultaneous Localization And Mapping}
\acro{MRA}{Multiresolution Analysis}
\acro{SDF}{Signed Distance Function}
\acro{TSDF}{Truncated Signed Distance Function}
\acro{ESDF}{Euclidean Signed Distance Function}
\acro{BoW}{Bag of Words}
\acro{MAV}{Micro Aerial Vehicle}
\acro{AABB}{Axis Aligned Bounding Box}
\acrodefplural{AABB}{Axis Aligned Bounding Boxes}
\acro{OBB}{Oriented Bounding Box}
\acro{ICP}{Iterative Closest Point}
\acro{SGD}{Stochastic Gradient Descent}
\acro{ATE}{Absolute Trajectory Error}
\acro{RMSE}{Root Mean Squared Error}
\acro{IMU}{Inertial Measurement Unit}
\acro{DOF}{Degrees of Freedom}
\acro{ROS}{Robot Operating System}
\acro{MRA}{Multi-Resolution Analysis}
\acro{FWT}{Fast Wavelet Transform}
\acro{ROC}{Receiver Operating Characteristic}
\acro{AUC}{Area under the ROC Curve}
\acro{TPR}{True Positive Rate}
\acro{FPR}{False Positive Rate}
\end{acronym}

\begin{abstract}

Volumetric maps are widely used in robotics due to their desirable properties in
applications such as path planning, exploration, and manipulation. Constant advances
in mapping technologies are needed to keep up with the improvements in sensor
technology, generating increasingly vast amounts of precise measurements. Handling
this data in a computationally and memory-efficient manner is paramount to
representing the environment at the desired scales and resolutions. In this work,
we express the desirable properties of a volumetric mapping framework through the
lens of multi-resolution analysis. This shows that wavelets are a natural foundation
for hierarchical and multi-resolution volumetric mapping. Based on this insight we design an efficient mapping system that uses wavelet decomposition. The
efficiency of the system enables the use of uncertainty-aware sensor
models, improving the quality of the maps. Experiments on both synthetic
and real-world data provide mapping accuracy and runtime performance comparisons
with state-of-the-art methods on both RGB-D and 3D LiDAR data. The framework is
open-sourced to allow the robotics community at large to explore this approach.

\end{abstract}

\IEEEpeerreviewmaketitle

\section{Introduction}

As robots move from tightly controlled spaces into our everyday lives, there is a growing need for them to autonomously navigate and work in increasingly large, unstructured, and unknown environments. For reliable deployments and robust operation over extended periods of time, robots need to build and maintain their representation of the world using only onboard sensing and computing. Doing this in a timely manner on compute restricted devices using sensors producing large amounts of data is a continual challenge in robotics.

Dense geometric environment representations are widely used to facilitate tasks ranging from navigation to inspection and manipulation, while also serving as building blocks for other representations. Robotics is a particularly challenging field for such representations, due to the demands placed on systems with limited computational resources. For example, building a map of an unknown environment while localizing in it with \ac{SLAM} requires the ability to update the map incrementally at interactive rates. To support high-level tasks such as exploration and navigation the representation must also differentiate between unknown space and observed (free or occupied) space. Finally, the map must be able to model arbitrary geometry with sufficient accuracy to guarantee safety when unexpected environmental structures or objects are encountered.

\begin{figure}
    \vspace{0.3em}
    \centering
    \includegraphics[width=0.485\textwidth]{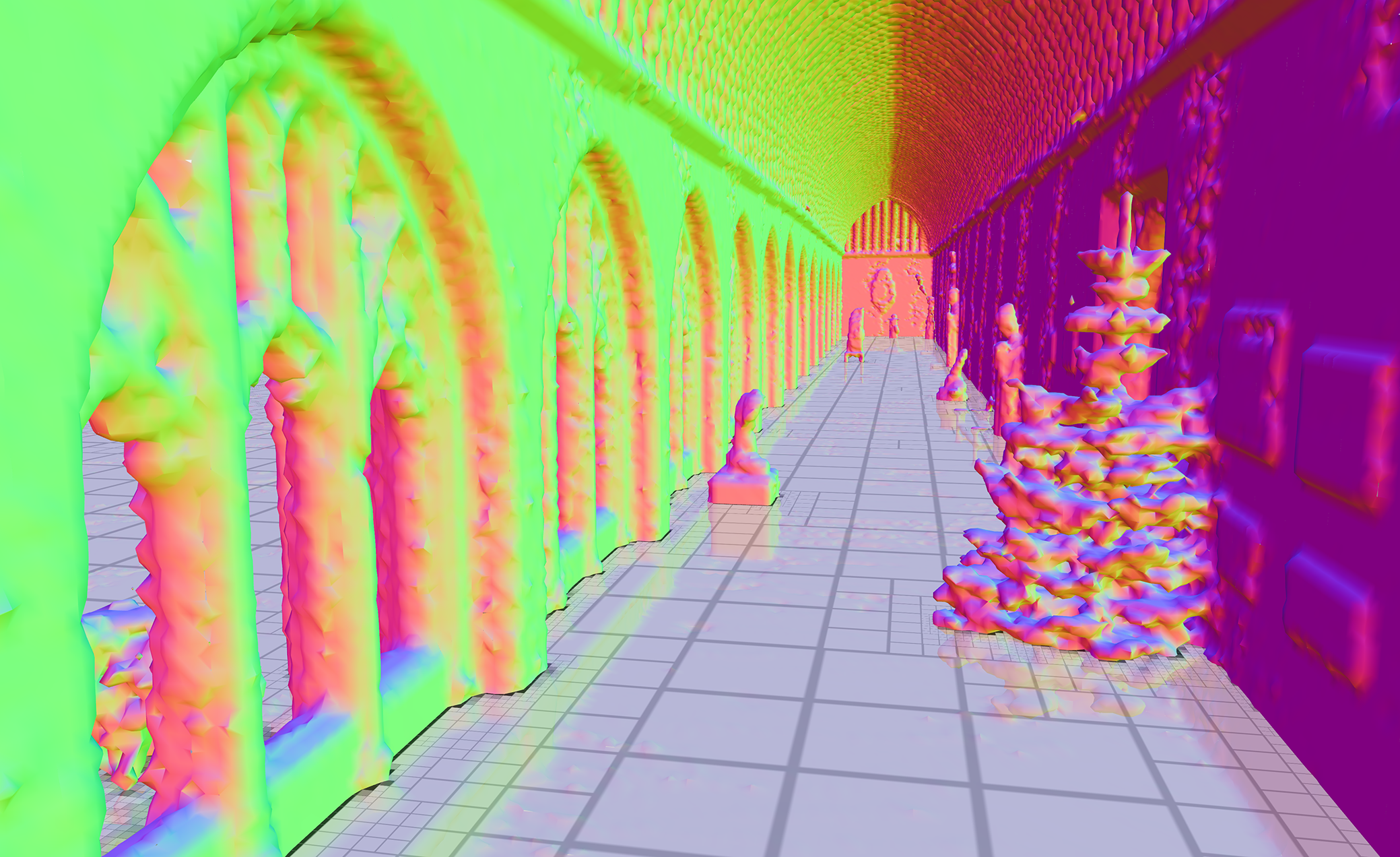}
    \caption{A reconstruction created by our proposed hierarchical volumetric mapping framework, \textit{wavemap}, highlighting its ability to accurately capture fine objects while also efficiently compressing free space as shown by the adaptive resolution along the transparent slice.}
    \label{fig:arche_reconstruction}
    \vspace{0.2em}
\end{figure}

% Memory
Volumetric map representations can be updated incrementally and explicitly represent unknown space. Furthermore, if a sufficiently high resolution is chosen, they can also represent object surfaces and unknown space boundaries of arbitrary topology. Beyond robotics, volumetric representations are commonly used in 3D reconstruction, reality capture, and augmented reality applications. However, a major drawback of volumetric representations is that their memory usage in naive implementations grows linearly with the observed volume and cubically with the chosen resolution. Several research efforts propose to use multi-resolution representations, often based on trees, and demonstrate significant improvements. In this work, we extend these efforts by approaching the problem from a formal signal processing and data compression perspective. Specifically, we propose to use wavelet compression to obtain a hierarchical volumetric representation. Using Haar wavelets we achieve state of the art lossless compression performance, while also allowing simple yet efficient updates and queries. This is achieved by compressing the occupancy information using a Haar wavelet decomposition and storing the individual decomposition components in a hierarchical data structure. The wavelet transform's linearity makes it possible to perform measurement updates directly in the map's compressed representation. Furthermore, when performing map updates we know that all resolution levels of the map are always up to date and in a valid state due to the Haar basis' orthogonality property. This obviates the need to perform maintenance operations or manual compression passes that are typically seen in other multi-resolution mapping frameworks.

% Accuracy
\newpage
Another trade-off made by many existing volumetric mapping methods is the reliance on simplified measurement models to achieve real-time update rates. A common approach is to use discrete occupancy updates, that systematically inflate obstacles and do not allow for surfaces to be reconstructed with sub-voxel accuracy. Measurement models based on \acp{TSDF} overcome the latter limitation but use a projective distance heuristic.
%From a theoretical point of view, this simplifying assumption makes it challenging to provide meaningful guarantees on the worst-case distance error and whether given elements will appear in the reconstruction at all. 
%From a practical point of view, \acp{TSDF} tend to miss thin, and potentially dangerous, obstacles such as tree branches, cables and fences.
Such approaches have a hard time reconstructing thin objects such as branches, cables, or fences. Furthermore, the implied assumption of 
%Furthermore, occupancy and \acp{TSDF} mapping frameworks typically neglect the thickness and angular uncertainty of rays.
infinitely thin rays, underlying these observation models, leads to aliasing artifacts in regions where the ray density is low compared to the voxel resolution. In addition to negatively affecting the reconstruction quality, the resulting high entropy regions are hard to compress. Besides alleviating the challenges mentioned above, modeling soft beams provides an opportunity to incorporate angular uncertainties from sensor calibration and pose estimation into the volumetric reconstruction process.
Thanks to the computational benefits of the Haar wavelet representation we can adopt a continuous occupancy measurement model, accounting for angular and range uncertainty, inspired by the work of \cite{loopClosedFormBayesianFusion2016}.

% Compute
In order to process data at sensor rate we introduce a specialized measurement integration algorithm that exploits a hierarchical measurement update approach with the information provided by the map itself. The proposed algorithm speeds up measurement integration while guaranteeing that the results are identical to a naive integrator applying the same measurement updates at the highest resolution throughout the field of view.

In summary, the main contribution of this paper is a volumetric mapping system that uses:
\begin{itemize}
    \item A wavelet-based hierarchical representation, that is guaranteed to keep the hierarchy consistent at all times;
    \item A continuous occupancy measurement model accounting for range and angular uncertainties;
    \item A highly-efficient coarse-to-fine measurement integrator that adapts to the observed structure;
\end{itemize}
The proposed framework is extensively evaluated on synthetic and real-world datasets with comparisons to several state-of-the-art methods. The results demonstrate that our approach is memory efficient yet produces high-quality maps, all while being computationally efficient. The entire framework is open-sourced\footnote{\url{https://github.com/ethz-asl/wavemap}} to enable the robotics community to build on these results.

\section{Related work}
\label{sec:related_work}

\subsection{Map model}

Two approaches are commonly used to represent maps in robotics \cite{cadenaPresentFutureSimultaneous2016}, sparse feature-based maps and dense maps. The first category uses sparse sets of distinctive features \cite{mur2015orb,schoenberger2016sfm} and excels at representing large environments but struggles to model the connectivity of surfaces and distinguish between free and unknown space. This makes it ideal for large scale mapping and localization tasks, but limits its use for manipulation, motion planning, and exploration tasks. The second paradigm uses a large number of geometric elements, such as as points \cite{engelLSDSLAMLargeScaleDirect2014,pomerleauComparingICPVariants2013}, surfels \cite{stuckler2014multi, behley2018efficient, henry2014rgb, park2018elastic,whelanElasticFusionDenseSLAM2015}, or meshes \cite{whelanKintinuousSpatiallyExtended2012} to model observed obstacles. Voxels, discretizing the space into squares or cubes of fixed size, are another common geometric primitive used to model both occupancy \cite{elfesUsingOccupancyGrids1989, hornungOctoMapEfficientProbabilistic2013} and signed distance information \cite{curlessVolumetricMethodBuilding1996,izadiKinectFusionRealtime3D2011, h.oleynikovaVoxbloxIncremental3D2017,kahlerVeryHighFrame2015}.

\subsection{Measurement model}

Approximations of the sensor's physical operation have been widely explored. Early approaches modeled uncertainties of the sensors explicitly \cite{elfesUsingOccupancyGrids1989}. Other approaches aim to achieve specific map properties, such as sharp map boundaries \cite{loopClosedFormBayesianFusion2016}.
However, when building 3D maps using precise sensors the computational cost incurred by these sensor models motivated the development of simpler ray-based models. These models treat observations as thin rays tracing through the world \cite{hornungOctoMapEfficientProbabilistic2013,h.oleynikovaVoxbloxIncremental3D2017}.
Machine-learning based methods exploit more complex relationships, such as inverse rendering \cite{mildenhallNeRFRepresentingScenes2020a,fridovich-keilPlenoxelsRadianceFields2022} or beam-to-beam interactions \cite{ocallaghan2011continuous}.

% To add: UFOMap
\subsection{Map storage}

The most common way to store volumetric maps is to discretize the space using a voxelgrid, i.e. a regular grid with fixed size voxels. In the beginning grids with a single fixed resolution \cite{elfesUsingOccupancyGrids1989,izadiKinectFusionRealtime3D2011} were used, but over time spatial data structures, such as hashed voxel blocks \cite{niessnerRealtime3DReconstruction2013}, trees \cite{hornungOctoMapEfficientProbabilistic2013}, or hybrids thereof \cite{vizzoVDBFusionFlexibleEfficient2022} were adopted. These structures fit the observed volume more tightly, can grow dynamically, and improve runtime.
To model expansive maps with varying levels of detail, multi-resolution maps \cite{hornungOctoMapEfficientProbabilistic2013,vespaEfficientOctreeBasedVolumetric2018,vespaAdaptiveResolutionOctreeBasedVolumetric2019,funkMultiResolution3DMapping2021,dubergUFOMapEfficientProbabilistic2020} are widely used due to being memory efficient and capable of adapting to the needed resolution. Many multi-resolution representations are also hierarchical, allowing users to query the map at varying resolutions  \cite{hornungOctoMapEfficientProbabilistic2013,funkMultiResolution3DMapping2021}.
Taking a signal processing perspective on compact map storage leads to the use of wavelet transforms \cite{manuelyguelWaveletOccupancyGrids2006}, which are inherently multi-resolution and hierarchical, or the discrete cosine transform \cite{schaeferDCTMapsCompact2018}. Recent learning-based methods, such as NERF \cite{mildenhallNeRFRepresentingScenes2020a} or occupancy prediction networks \cite{mescheder2019occupancy, peng2020convolutional}, take a different approach and learn the coefficients of a neural network that predicts map information at arbitrary coordinates.

% To add: UFOMap, OpenChisel
\subsection{Map updates}

The manner in which maps are updated with new observations is crucial for the efficiency and map quality. Early volumetric frameworks evaluated the measurement model for all voxels in the observed volume \cite{elfesUsingOccupancyGrids1989,izadiKinectFusionRealtime3D2011}. This was improved by tracing rays from the sensor's center to each measured point and updating the voxels that are intersected by the ray \cite{hornungOctoMapEfficientProbabilistic2013,h.oleynikovaVoxbloxIncremental3D2017}. 
Advances in sensor technology, enabling high resolution maps spurred further efficiency improvements, such as ray-tracers that bundle \cite{h.oleynikovaVoxbloxIncremental3D2017} or sub-sample \cite{millaneCbloxScalableConsistent2018,dubergUFOMapEfficientProbabilistic2020} similar rays, or rate limit voxel updates \cite{millaneCbloxScalableConsistent2018}. While efficient, these integrators can produce ``holes'' in the map depending on the sensor's ray density. This motivates the use of projective integrators which avoid this problem by interpolating the depth image \cite{curlessVolumetricMethodBuilding1996,loopClosedFormBayesianFusion2016}. Other approaches to avoid resolution-related issues include multi-resolution integrators, ray-tracing \cite{dubergUFOMapEfficientProbabilistic2020} or projective \cite{vespaAdaptiveResolutionOctreeBasedVolumetric2019},  which reduce the update resolution with distance, as well as methods analyzing the measurement update regularity  \cite{fuhrmannFusionDepthMaps2011,funkMultiResolution3DMapping2021}.
While efficient, hierarchical volumetric maps require maintenance to keep the information in the different levels coherent. Octomap \cite{hornungOctoMapEfficientProbabilistic2013} employs a fine-to-coarse scheme, integrating measurements at the finest resolution and synchronizing coarser levels in a maintenance pass. Supereight \cite{vespaAdaptiveResolutionOctreeBasedVolumetric2019,funkMultiResolution3DMapping2021} performs multi-resolution updates and synchronizes the remaining levels using an upward and downward propagation scheme.

In contrast to others, our volumetric ray-tracing method uses a wavelet decomposition-based representation which implicitly synchronizes all hierarchy levels at once. Additionally, unlike most ray-tracing methods we use a continuous sensor model, taking angular uncertainty into account, to improve map accuracy.
\section{Multi-Resolution Analysis and Wavelets}
\label{sec:mra}

Multi-resolution representations have been the subject of intensive study by communities ranging from computer vision \cite{burtLaplacianPyramidCompact1983} to physics and mathematics \cite{daubechiesTenLecturesWavelets1992,mallatWaveletTourSignal2009}. Mallat and Meyer formalized the expected properties of multi-resolution representations as the \ac{MRA} conditions \cite{mallatWaveletTourSignal2009}. The full \ac{MRA} conditions are summarized in appendix \ref{appendix:mra}. In informal terms, they state that increasing the resolution should only add detail and eventually make it possible to represent any signal. Two further requirements are self-similarity in space and in scale. In a mapping context, these imply that the map should behave the same regardless of our frame of reference and choice of units.

A corollary of the fact that increasing the resolution only adds information is that, in areas that are stored at multiple resolutions, the lower resolutions do not carry any unique information and storing them explicitly is redundant. This motivates the use of wavelet decompositions, which allow us to work with maps that form valid \acp{MRA} while only storing and processing the differences between the resolution levels.
A given wavelet decomposition is characterized by its chosen scaling function and complementary wavelet function. In this work, we focus on the Haar wavelet and scaling function, which form an orthogonal basis. A summary of orthogonal wavelet bases is provided in appendix \ref{appendix:orthogonal_wavelet_bases}. This orthogonality is particularly beneficial because it guarantees that any given volumetric map is characterized by a unique combination of wavelet coefficients. Thus, there are no redundant coefficients that can go out of sync and manually have to be updated after integrating new measurements. Another interesting property of Haar wavelets is that the basis resulting from its scaling functions correspond to box functions arranged to span the cells of a regular grid. Therefore, Haar decompositions can represent anything a regular grid map can, while bringing significant benefits in terms of compression and implicitly maintaining the hierarchy's consistency.

\section{Method}
In the following, we describe the components of our approach. We first explain how the map's occupancy posterior can be efficiently updated in its compressed state, thanks to the properties of the wavelet transform. Next, we derive our continuous sensor model, which captures range and angular uncertainties associated with the measurements. After that, we derive an error bound which enables early stopping during the coarse-to-fine observation integration process. Further performance improvements are obtained by skipping updates that do not change the state of the map. Finally, we illustrate how all these pieces fit together with an algorithmic overview.

\subsection{Measurement integration}

In the following we will explain how the use of wavelets enables efficient measurement integration. As each new beam endpoint measurement $\vec{z}$ arrives, the map's Bayesian occupancy posterior $p(m_{\vec{x}}|\vec{z}_{1:t})$, estimated at each point $\vec{x}$ in the map $m$, can incrementally be updated using
\begin{equation}
    \mathcal{L}_p(m_{\vec{x}}|\vec{z}_{1:t}) = \mathcal{L}_p(m_{\vec{x}}|\vec{z}_{1:t-1}) + \mathcal{L}_s(m_{\vec{x}}|\vec{z}_t),
\end{equation}
where $s(m_{\vec{x}}|\vec{z}_t)$ is the sensor's inverse measurement model and the log-odds formulation, $\mathcal{L}_p = \log\frac{p}{1-p}$, is used to make the update linear. As the wavelet transform $\mathcal{W}$ is also linear, the update equation for all cells in the map becomes:
\begin{equation}
    \mathcal{W}\left(\mathcal{L}_p(m|\vec{z}_{1:t})\right) =
        \mathcal{W}\left(\mathcal{L}_p(m|\vec{z}_{1:t-1})\right) +
        \mathcal{W}\left(\mathcal{L}_s(m|\vec{z}_t)\right).
\end{equation}
Therefore, once the compressed measurement update $\mathcal{W}\left(\mathcal{L}_s(m|\vec{z}_t)\right)$ is computed, the map can be updated directly in wavelet space.
This is avoids the costly process, employed by other methods, of decompressing the map's observed area, applying the update, and compressing the map again. Computing $\mathcal{W}\left(\mathcal{L}_s(m|\vec{z}_t)\right)$ is efficient thanks to the \ac{FWT} (Appendix \ref{appendix:fast_wavelet_transform}), which is typically initialized by computing the orthogonal projection of the original signal onto the scaling functions at a pre-determined finest resolution. 

Since the wavelet transform itself is lossless, the reconstruction error is fully determined by how well the initial \ac{FWT} projection approximates the original update.
Most applications use a constant finest resolution, but this is not mandatory. Given that inverse sensor models tend to be smooth throughout most of the observed volume, only raising the resolution close to surfaces would improve efficiency and the maximum achievable detail.

\subsection{Measurement models}
\label{sec:measurement_model}

\begin{figure*}[bt]
    \centering
    \subfloat[]{\frame{\includegraphics[width=0.25\textwidth]{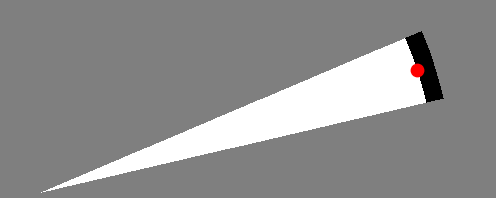}}%
    \label{fig:occupancy_given_gt}}
    \hfil
    \subfloat[]{\frame{\includegraphics[width=0.25\textwidth]{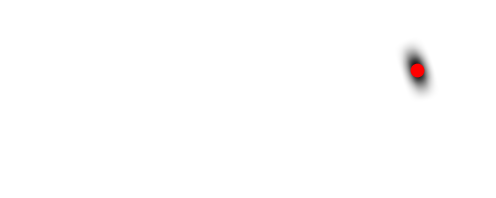}}%
    \label{fig:gt_given_measurement}}
    \hfil
    \subfloat[]{\frame{\includegraphics[width=0.25\textwidth]{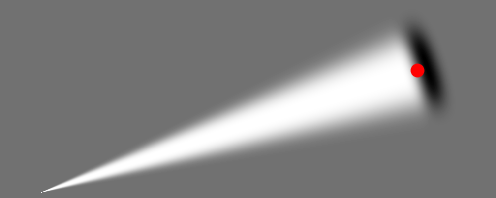}}%
    \label{fig:inverse_measurement_model}}
    \hfil
    \subfloat[]{\frame{\includegraphics[width=0.25\textwidth]{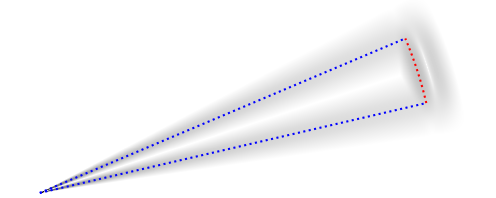}}%
    \label{fig:inverse_measurement_model_local_maxima}}
    \caption{Figure illustrating our proposed models for a) the occupancy given the true beam endpoint $\bar s(m_\vec{x}|\bar{\vec{z}})$ (eq. \ref{eq:gt_surface_model}), b) the position of the true endpoint given a noisy measurement $o(\bar{\vec{z}}|\vec{z})$ (eq. \ref{eq:gt_point_given_measurement}), c) the complete inverse measurement model $s(m_\vec{x}|\vec{z})$ (eq. \ref{eq:concrete_inverse_occupancy_meas_model}), and d) the local maxima used to derive the worst-case error bounds. Values of 0.0, 0.5 and 1.0 are shown in white, grey and black, respectively. The true beam endpoint is indicated in red. Uncertainties are exaggerated for illustration.}
    \label{fig:beam_model_derivation}
    % \vspace{-1em}
\end{figure*}

In order to derive multi-resolution sampling and integration approaches, it is important that the chosen inverse measurement model $s(m_{\vec{x}}|\vec{z}_t)$ is well-defined at all points $\vec{x}$ in the observed volume. We propose to extend the continuous occupancy model introduced in \cite{loopClosedFormBayesianFusion2016} by modeling the angular uncertainty of each measured beam, in addition to range uncertainty.
We model the probability of occupancy $s(m_\vec{x} | \vec{z})$ at a point $\vec{x}$ for a single beam $\vec{z}$ by correlating the probability of occupancy given the beam's true endpoint $\bar s(m_\vec{x} | \bar{\vec{z}})$ with the distribution of the true endpoint position given a noisy observation $o(\bar{\vec{z}} | \vec{z})$, i.e.:
\begin{equation}
    \label{eq:general_inverse_occupancy_meas_model}
    s(m_\vec{x} | \vec{z}) = \int_{\mathbb{S}} \bar s(m_\vec{x} | \bar{\vec{z}}) o(\bar{\vec{z}} | \vec{z}) d\bar{\vec{z}},
\end{equation}
where $\vec{x}$, $\bar{\vec{z}}$, and $\vec{z}$ are expressed in sensor coordinate space $\mathbb{S}$, and the beam start point is at its origin.
Extending \cite{loopClosedFormBayesianFusion2016} to include angular uncertainty, we define $\bar s(m_\vec{x} | \bar{\vec{z}})$ as
\begin{align}
    \label{eq:gt_surface_model}
    \bar s(m_\vec{x} | \bar{\vec{z}}) &= \bar s(m_\vec{x} | \bar{z}_r, \bar{z}_\theta) \nonumber \\
    & = 
    \begin{cases}
        0           & x_r < \bar{z}_r                           \ \wedge\ | x_\theta - \bar{z}_\theta | \leq \tau_\theta \\
        1           & \bar{z}_r \leq x_r \leq \bar{z}_r + \tau_r    \ \wedge\ | x_\theta - \bar{z}_\theta | \leq \tau_\theta \\
        \frac{1}{2} & \text{otherwise}
    \end{cases}
\end{align}
where $\vec{\tau}$ is an assumed surface thickness parameter in sensor coordinates, see Fig. \ref{fig:occupancy_given_gt} for a visualization. The subscript $r$ refers to the axis perpendicular to the sensor's image plane, whereas $\theta$ refers to the offset along the image plane\footnote{For pinhole camera projection models, $r$ corresponds to the depth coordinate and $\theta$ to the reprojection error. For spherical projection models, e.g. certain LiDARs, $r$ refers to the range coordinate and $\theta$ to the relative angle.}.

Our model assumes that the noise on the measurement beam endpoint position is normally distributed in sensor coordinates, as
\begin{equation}
    o(\vec{z} | \bar{\vec{z}}) \sim \mathcal{N}(\bar{\vec{z}}, \Sigma),
    \label{eq:obs-noise}
\end{equation}
where $\Sigma$ is the measurement noise covariance matrix.
If $\Sigma$ is diagonal and $\bar{\vec{z}}$ has a uniform prior, the $r$ and $\theta$ components are independent and eq. \ref{eq:obs-noise} can be simplified as follows:
\begin{equation}
    o(\bar{\vec{z}} | \vec{z})
    = o(z_r | \bar{z}_r) o(z_\theta | \bar{z}_\theta)
    = \mathcal{N}(\bar{z}_r, \sigma_r) \mathcal{N}(\bar{z}_\theta, \sigma_\theta).
\end{equation}
We approximate the normal distributions with quadratic B-splines, as in \cite{loopClosedFormBayesianFusion2016}, such that $o(z | \bar{z}) \simeq q(\frac{\bar{z}-z}{\sigma})$, where
\begin{equation}
    q(t) =
    \begin{cases}
        \frac{1}{16} (3 + t)^2 & -3 \leq t \leq -1 \\
        \frac{1}{8} (3 - t^2)  & -1 < t < 1 \\
        \frac{1}{16} (3 - t)^2 & \hphantom{-} 1 \leq t \leq 3 \\
        0                      & \hphantom{-} \text{otherwise} \\
    \end{cases}.
\end{equation}
The distribution of the true beam endpoint position given a noisy measurement (Fig. \ref{fig:gt_given_measurement}) then becomes:
\begin{equation}
    \label{eq:gt_point_given_measurement}
    o(\bar{\vec{z}} | \vec{z}) = q\left(\frac{z_r - \bar{z}_r}{\sigma_r}\right) q\left(\frac{z_\theta - \bar{z}_\theta }{\sigma_\theta}\right).
\end{equation}
As motivated in \cite{loopClosedFormBayesianFusion2016}, we match the surface thicknesses to half the width of their respective B-splines, i.e. $\tau_r = 3\sigma_r$ and $\tau_\theta = 3\sigma_\theta$. This ensures that the measurement model is continuous and that $\mathcal{L}_p(m_\vec{x}|\vec{z}_{1:t})$ converges to $0$ if $\vec{x}$ lies on an object's surface.

After substituting \ref{eq:gt_surface_model} and \ref{eq:gt_point_given_measurement} into \ref{eq:general_inverse_occupancy_meas_model}, the full inverse measurement model (Fig. \ref{fig:inverse_measurement_model}) becomes:
\begin{multline}
\label{eq:concrete_inverse_occupancy_meas_model}
    s(m_\vec{x}|\vec{z}) = \int_0^\infty \int_{-\infty}^{\infty} \bar s(m_\vec{x} | \bar{z}_r, \bar{z}_\theta)q(v) q(w)d\bar{z}_\theta\,d\bar{z}_r \\
    = \frac{1}{2} + \left(Q(v) - \frac{Q(v-3)}{2} - \frac{1}{2}\right)\Big(Q(w+3) - Q(w-3) \Big),
\end{multline}
where $Q(t)$ refers to the cumulative distribution of $q(t)$, i.e. the cubic B-splines resulting from $Q(t) = \int_{-\infty}^{t} q(u) du$, and $v = \frac{z_r-\bar{z}_r}{\sigma_r}$, $w = \frac{z_\theta-\bar{z}_\theta}{\sigma_\theta}$.

For depth cameras, the depth uncertainty is often set as $\sigma_r(x) = \kappa_r x_r^2$, where $\kappa_r$ depends on the sensor setup and post-processing algorithms. For laser-based sensors, the range error is usually assumed to not vary with range, thus $\sigma_r = \kappa_r$ where $\kappa_r$ is indicated on the sensor's datasheet.

\subsection{Worst-case update error bounds}
\label{sec:mra_error_bounds}

From the MRA theory (Section \ref{sec:mra}) we know that at some point, integrating information at finer levels of the hierarchy no longer improves the representation. Therefore, to fully exploit the coarse-to-fine measurement integration scheme of our method we need to know at what level of the hierarchy we can stop integrating data. This requires determining, for each point $\vec x$, the resolution beyond which no further improvements are possible, which we achieve by deriving a conservative approximation error bound. As this work focuses on the use of Haar wavelets, we can exploit a property unique to them, namely that neighbors at the same resolution do not overlap. This results in the leaves of our multi-resolution Haar decomposition perfectly partitioning the original space into non-overlapping cubes of varying sizes.
Since Haar scaling functions are constant over their support, the worst-case error $\epsilon_\text{max}$ within each space partition, or voxel, $\mathcal{V}$ is given by:
\begin{equation}
    \epsilon_\text{max}(\mathcal{L}_s(m, \vec{z}), \mathcal{V}) = \max_{\vec{x}\in\mathcal{V}} \left| \mathcal{L}_s(m_{\vec{x}'}, \vec{z}) - \mathcal{L}_s(m_\vec{x}, \vec{z}) \right|,
\end{equation}
where $\vec{x}'$ is the chosen sample point, which we set to be the partition's center.

Since $\epsilon_\text{max}$ has to be evaluated millions of times per second in practice, we simplify the computation by only considering three cases based on the state of the space partition, defined as follows:
\begin{multline}
    \label{eq:update_type}
    \text{update\_type}(\mathcal{V}, \vec{z}_t) = \\
    \begin{cases}
        \text{FullyUnobserved}     &  \forall \vec{x}\in\mathcal{V}: \mathcal{L}_s(m_\vec{x}|\vec{z}_t) = 0\\
        \text{PossiblyOccupied}    &  \exists \vec{x}\in\mathcal{V}: \mathcal{L}_s(m_\vec{x}|\vec{z}_t) > 0\\
        \text{FreeOrUnobserved}    & \text{otherwise}
    \end{cases}
\end{multline}
Looking at Eq. \ref{eq:concrete_inverse_occupancy_meas_model} we can see that the gradient of $s(m_\vec{x}|\vec{z})$ is zero in FullyUnobserved areas and reaches local maxima where $x_\theta = z_\theta \pm 3 \sigma_\theta$ or $x_r = z_r$ as illustrated in Fig. \ref{fig:inverse_measurement_model_local_maxima}. Using the fact that 
\begin{equation}
    \frac{\partial s(m_\vec{x}|\vec{z})}{\partial x_\theta}
        \bigg\rvert_{x_\theta = z_\theta \pm 3 \sigma_\theta} =
        \frac{3}{16 \sigma_\theta}, \quad
    \frac{\partial s(m_\vec{x}|\vec{z})}{\partial x_r}
        \bigg\rvert_{x_r = z_r} =
        \frac{3}{8 \sigma_r}
\end{equation}
and assuming the worst-case orientations for a cube-shaped partition $\mathcal{V}$, i.e. its diagonal projected into sensor coordinates $r$ and $\theta$ aligns with either gradient, we obtain the following bounds for the approximation error for the three cases:
\begin{equation}
\label{eq:worst_case_approximation_error}
\epsilon_\text{max}(\mathcal{V}) =
    \begin{cases}
        0 & \text{FullyUnobserved}\\
        \text{max}\left(\frac{3\mathcal{V}_{h_\theta}}{16 \sigma_\theta}, \frac{3\mathcal{V}_{h_r}}{8 \sigma_r}\right) & \text{PossiblyOccupied}\\
        \frac{3\mathcal{V}_{h_\theta}}{16 \sigma_\theta} & \text{FreeOrUnobserved}
    \end{cases}
\end{equation}
where $\mathcal{V}_h$ is the maximum distance a sample can have to $\mathcal V$'s center, namely half of $\mathcal{V}$'s diagonal. Note that $\mathcal{V}_{h_\theta}$ decays quickly as the distance to the sensor increases.

\subsection{Saturated region skipping}

To preserve the ability to quickly adapt the map when dynamic parts of the environment change, we impose upper and lower bounds on the occupancy posterior $\mathcal{L}_p(m_\vec{x}|\vec{z}_{1:t})$, as proposed by \citet{yguelUpdatePolicyDense2008}. As observed by \citet{hornungOctoMapEfficientProbabilistic2013}, this clamping policy also significantly improves compression performance by encouraging the majority of the map's posterior to converge to constant values. Namely to the lower bound in areas that are consistently observed as being free, and to the upper bound in areas that are consistently observed as being occupied.
We propose to exploit this saturating behavior further to reduce the computational cost of map updates. Applying negative occupancy (free-space) updates in areas where the posterior has already reached the lower bound has no effect, as the updates are canceled out by the clamping operation. Similarly, the posterior is not affected by skipping positive occupancy updates in areas that already converged to the upper bound. Skipping saturated regions leads to a particularly high speedup if it can be done in a coarse-to-fine manner, but doing so is only safe if the map's lower resolutions are always up to date. Both properties are met by our representation and integration scheme. An algorithm that interleaves saturated region skipping, adaptive sampling, and thresholding will be discussed in the next section.

\subsection{Algorithm and data structure}
\label{subsec:algorithm_and_datastructure}
As described previously, Haar scaling functions do not overlap with their neighbors at the same resolution and perfectly partition the space. The support of the scaling functions in a multi-resolution Haar decomposition is, therefore, identical to the hierarchical partitioning scheme of octrees. We can thus store the wavelet coefficients in any optimized octree data structure that allows data to be attached to both inner and leaf nodes, such as supereight \cite{vespaEfficientOctreeBasedVolumetric2018} or OpenVDB \cite{musethOpenVDBOpensourceData2013}.

Leveraging the idea that increasing the resolution in \acp{MRA} only adds information, our proposed adaptive multi-resolution update algorithm determines the appropriate update resolution for all points in the observed volume in a coarse-to-fine manner (Section \ref{sec:mra_error_bounds}). The algorithm is initialized at the octree's root and recursively evaluates its children, as illustrated in Algorithm~\ref{algo:recursive_adaptive_sampler}. Each recursive call starts by checking which of the three possible update cases, eq. \ref{eq:update_type}, applies to the current node's partition $\mathcal{V}$. If no parts of the partition have been observed by the current measurement $\vec{z}_t$, or if saturated region skipping applies, no updates are needed. Otherwise, we continue by checking if the  approximation error at the partition's current resolution is acceptable. If this is the case, we evaluate the inverse measurement model $s(m_{\vec{x}'}|\vec{z}_t)$ at the partition's center and integrate the update into the map. If none of the previous criteria were met, a higher resolution is needed and the recursive function is called for each of the octree node's sub-divisions (octants).
In practice, we also compress the measurement update using the wavelet transform and need to traverse the map's data structure. Both of these operations can efficiently be interleaved with the recursive adaptive sampling procedure. Note that although the presented algorithm is recursive, great flexibility exists for its implementation. For example, since each Haar scaling function only overlaps with its parent and children, all partitions at a given resolution and their descendants can be updated in parallel.

\begin{algorithm}[bt]
    \SetAlgoLined
    \SetNoFillComment
    \DontPrintSemicolon

    \KwIn{%
        Current measurement $\vec z_t$,\;
        \Indp\Indp Previous map posterior $p(m \mid \vec z_{1:t-1})$,\;
        Lower log-odds threshold $\mathcal L_\text{min}$,\;
        Approximation error threshold $\epsilon_\text{thresh}$,\;
        Maximum resolution $\text{res}_\text{max}$,\;
        Octree's root partition $\mathcal V_\text{root}$
    }
    \KwOut{Updated map posterior $p(m \mid \vec z_{1:t})$}

    \SetKwProg{Fn}{Function}{ is}{end}

    \SetKwFunction{updateType}{UpdateType}
    \SetKwFunction{approxError}{ApproximationError}
    \SetKwFunction{recursiveAdaptiveUpdate}{RecursiveAdaptiveUpdate}
    
    \SetKwData{UT}{update\_type}
    \SetKwData{FU}{FullyUnobserved}
    \SetKwData{FOU}{FreeOrUnobserved}

    \Fn{\recursiveAdaptiveUpdate{$\mathcal{V}, \vec z_t$}}{
        \tcp{Use Eq.\ref{eq:update_type} to skip partitions}
        $\UT \gets \updateType(\mathcal{V}, \vec z_t)$\;
    
        \If{$\UT == \FU$}{%
            \Return
        }
        \If{
            $(\UT == \FOU$ \\
            \mbox{}\phantom{\textbf{if} \itshape(}$\mathbf{\;and\;} \mathcal{L}_p(m_{\mathcal{V}} \mid z_{1:t-1}) \leq \mathcal{L}_\text{min})$
        }{%
            \Return
        }
        \tcp{Use Eq.\ref{eq:worst_case_approximation_error} to terminate early}
        $\epsilon_\text{max}(\mathcal{V}) \gets \approxError(\mathcal{V}, \vec z_t)$\;
        \If{$(\mathcal{V}_\text{res} == \text{res}_\text{max} \mathbf{\;or\;} \epsilon_\text{max}(\mathcal{V}) < \epsilon_\text{thresh})$
        }{%
            $\mathcal{L}_p(m_{\mathcal{V}} \mid \vec z_{1:t}) \gets
                \mathcal{L}_p(m_{\mathcal{V}} \mid \vec z_{1:t-1})$\\
                \mbox{}\phantom{$\mathcal{L}_p(m_{\mathcal{V}} \mid \vec z_{1:t}) \gets$}
                $+\mathcal{L}_s(m_{\mathcal{V}} \mid \vec z_t)$\;
            \Return
        }
        \tcp{Otherwise, increase resolution}
        \For{$\mathcal{V}_\text{child} \in \mathcal{V}$}{%
            $\recursiveAdaptiveUpdate(\mathcal{V}_\text{child}, \vec z_t)$\;
        }
    }

    \tcp{Initialize map and start recursion}
    $p(m \mid \vec z_{1:t}) \gets p(m \mid \vec z_{1:t-1})$\;
    $\recursiveAdaptiveUpdate(\mathcal{V}_\text{root}, \vec z_t)$\;

    \caption{Wavemap recursive update}
    \label{algo:recursive_adaptive_sampler}
\end{algorithm}
\section{Experiments}

We evaluate our approach on three different datasets, featuring depth cameras and LiDARs, in indoor as well as outdoor environments.
Comparisons are presented to three state-of-the-art volumetric mapping frameworks: octomap \cite{hornungOctoMapEfficientProbabilistic2013}, voxblox \cite{h.oleynikovaVoxbloxIncremental3D2017}, and supereight2 \cite{funkMultiResolution3DMapping2021}. Octomap and supereight2 are both used in multi-resolution occupancy mapping-mode. Voxblox only supports \acp{TSDF} mapping and is configured to use its default `fast' integration method. In terms of implementation details, all approaches are evaluated using their publicly available reference implementations\footnote{\url{https://github.com/OctoMap/octomap_mapping}}\footnote{\url{https://bitbucket.org/smartroboticslab/supereight2}}\footnote{\url{https://github.com/ethz-asl/voxblox}} and wrapped with the same code to process the training data.

For each dataset, we split the original data into training and test sets by reserving every 20th observation for testing and use the remaining frames for mapping. Test points are generated by sampling points along all rays in each test observation, with points along the beam being in free space and the endpoint being occupied. To obtain insights into the behavior of the different methods in various scenarios we compute the distance of each free-space test point to the closest surface point. This allows us to evaluate the performance in different range bands, including: i) small negative values assessing the ability to capture thin objects, ii) distances close to zero to evaluate the surface reconstruction quality, and iii) larger distances to obstacles to detect possible biases or approximation errors. This approach also avoids diluting a small number of challenging situations with a large number of easy-to-classify free space observations.

For each experiment, we report the overall \ac{AUC} as a general indicator of classification performance. By integrating the \ac{ROC} curve, the \ac{AUC} quantifies how well each classifier discriminates free and occupied space regardless of the classification threshold. We also report the classification accuracy for the individual range bands. Note that different accuracies can be obtained based on the chosen classification threshold. For this study, we set the thresholds for each framework on each dataset to the value that maximizes the difference between the \ac{TPR} and the \ac{FPR}, weighed equally.

\begin{table}[bt]
    \caption{Area Under the ROC Curve results for both datasets. Higher is better. The corresponding resource usages are in table \ref{tab:comp_perf}.}
    \label{table:auc}
    \begin{tabular}{llccccc}
        \hline
        \multirow{ 2}{*}{Dataset} & \multirow{ 2}{*}{Res} & \multirow{ 2}{*}{octomap} & super- & \multirow{ 2}{*}{voxblox} & \textit{ours} & \textit{ours} \\
         & & & eight2 & & \textit{(rays)} & \textit{(beams)} \\\hline
        Panoptic & 5cm & 0.95 & 0.93 & \textbf{0.99} & \textbf{0.99} & \textbf{0.99} \\
        Flat & 2cm & 0.99 & 0.95 & \textbf{1.00} & \textbf{1.00} & \textbf{1.00} \\ \hline
        Newer & 20cm & 0.82 & 0.87 & \textbf{0.92} & 0.91 & 0.91 \\
        College & 5cm & 0.90 & 0.89 & \textbf{0.97} & 0.94 & \textbf{0.97} \\ \hline
    \end{tabular}
    \vspace{-0.4em}
\end{table}

\subsection{Accuracy evaluations}

\subsubsection{Panoptic mapping dataset}

\begin{figure}[bt]
    \centering
    \includegraphics[width=0.475\textwidth]{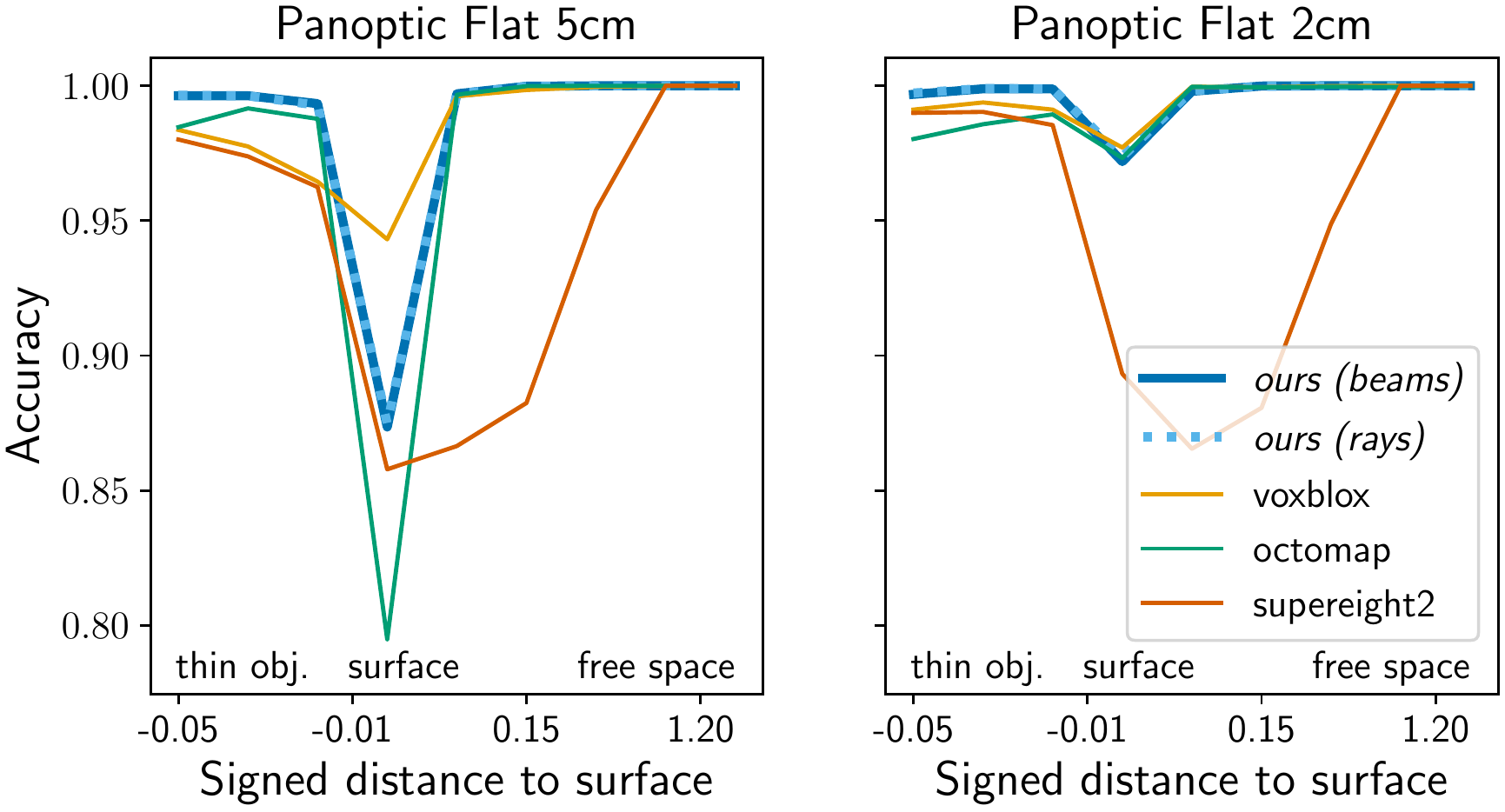}
    \caption{Accuracy in function of the distance to the surface on the Panoptic mapping dataset at different resolutions. Higher is better.}
    \label{fig:panoptic_flat_quantitative}
    \vspace{-1em}
\end{figure}

The first set of experiments is conducted using ``Run 1" of the panoptic mapping dataset \cite{schmidPanopticMultiTSDFsFlexible2022}, which features depth camera recordings of a simulated studio apartment including realistic household objects. Octomap and voxblox do not support depth images directly, and hence the dataset's images were first converted to pointclouds using the pinhole projection model used by both supereight2 and our method. The camera poses were obtained from the ground truth.

\begin{figure*}[bt]
\setlength\tabcolsep{0pt}
\renewcommand{\arraystretch}{0}
\centering
\begin{tabular}{cccc}
 \includegraphics[width=0.25\textwidth]{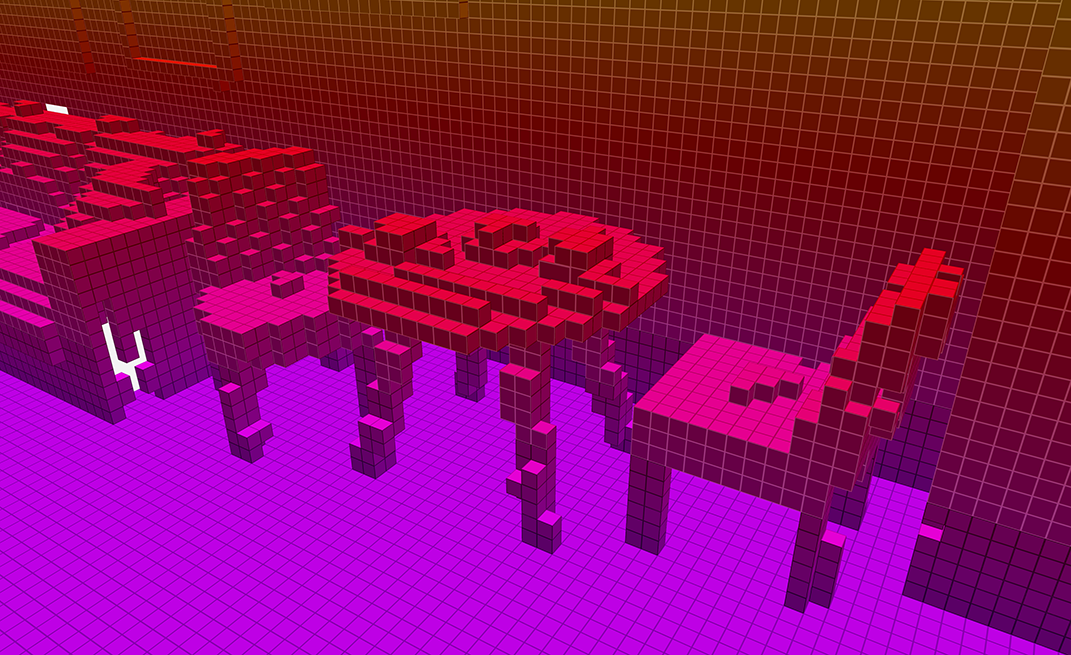} & 
 \includegraphics[width=0.25\textwidth]{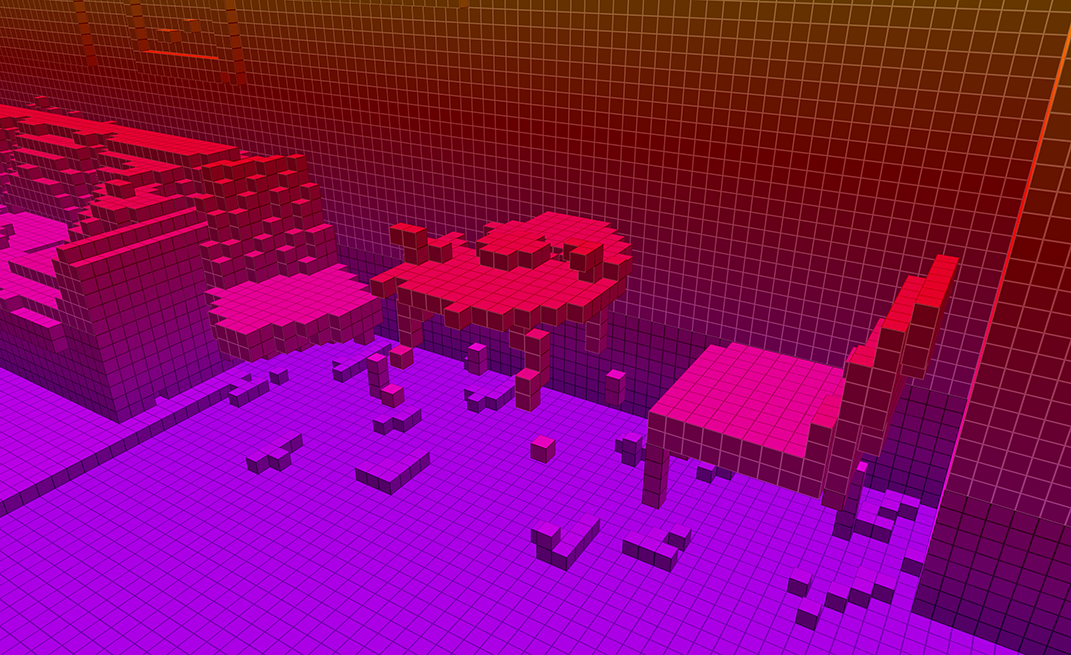} & 
 \includegraphics[width=0.25\textwidth]{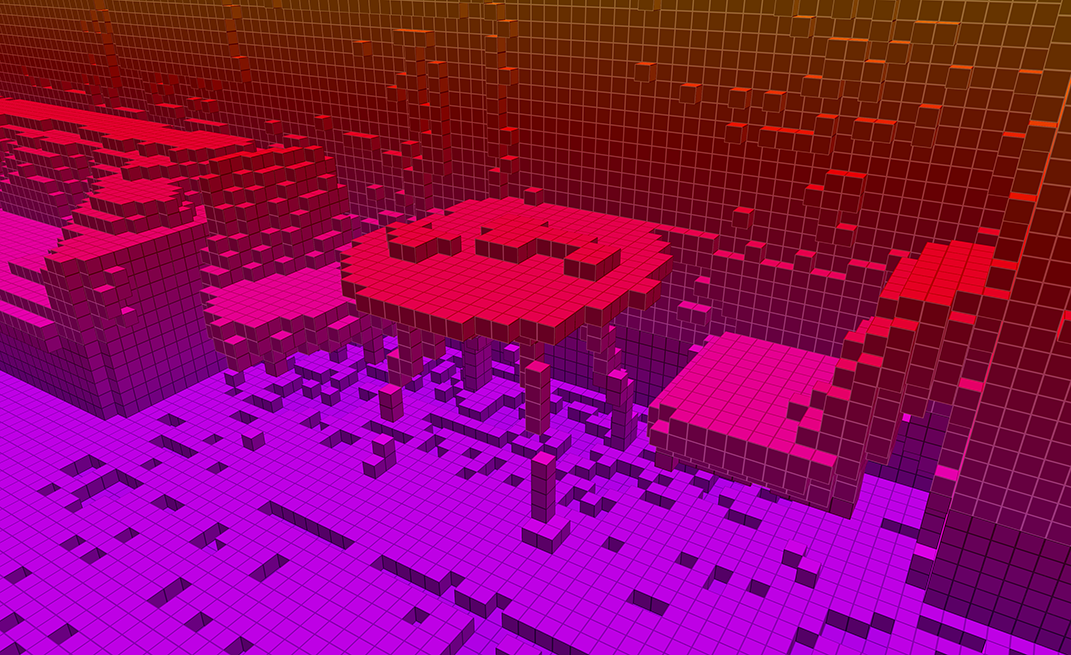} & 
 \includegraphics[width=0.25\textwidth]{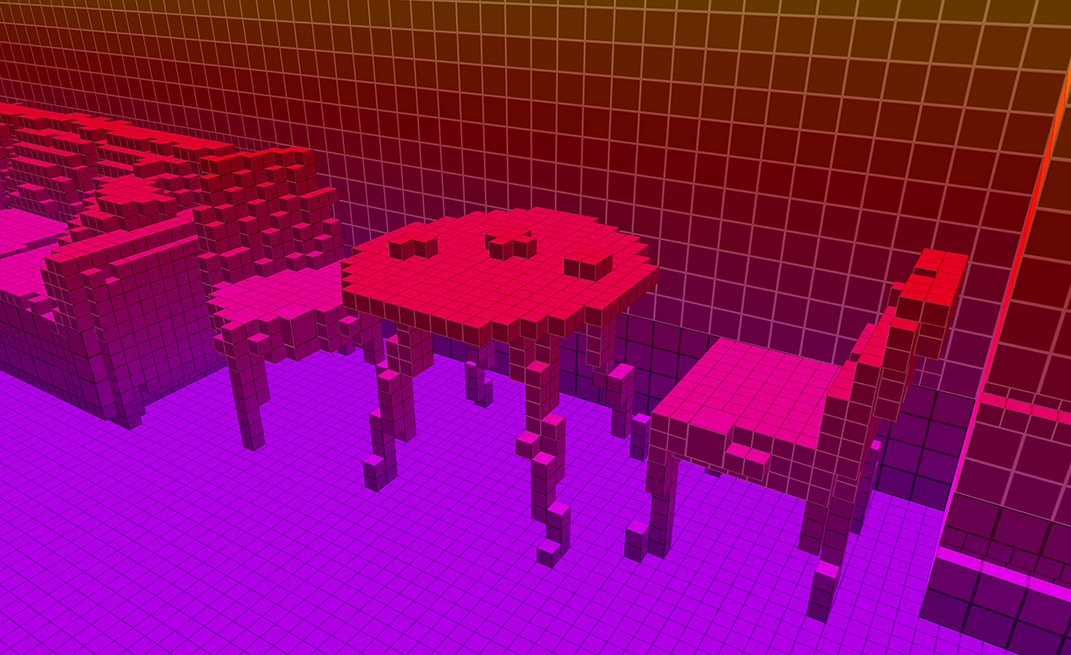} \\
 \includegraphics[width=0.25\textwidth]{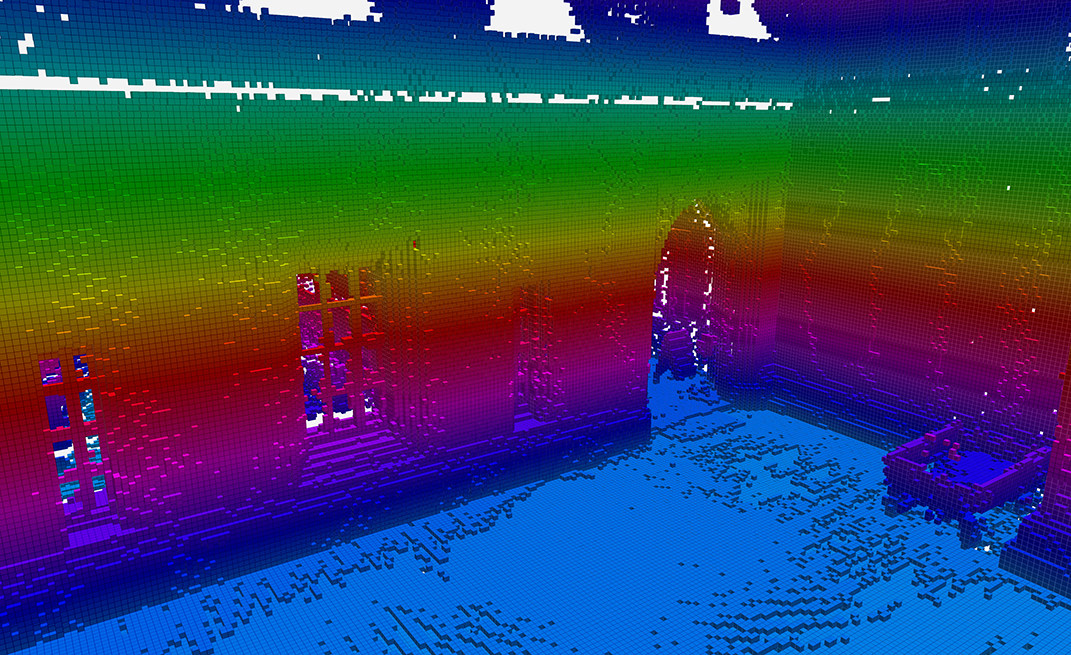} & 
 \includegraphics[width=0.25\textwidth]{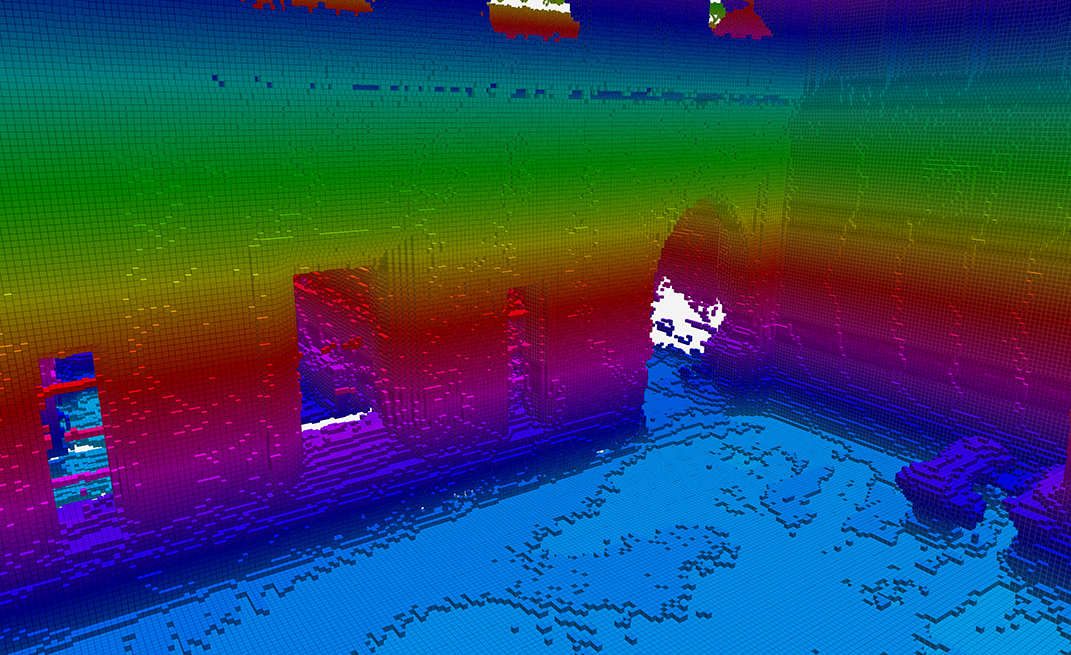} & 
 \includegraphics[width=0.25\textwidth]{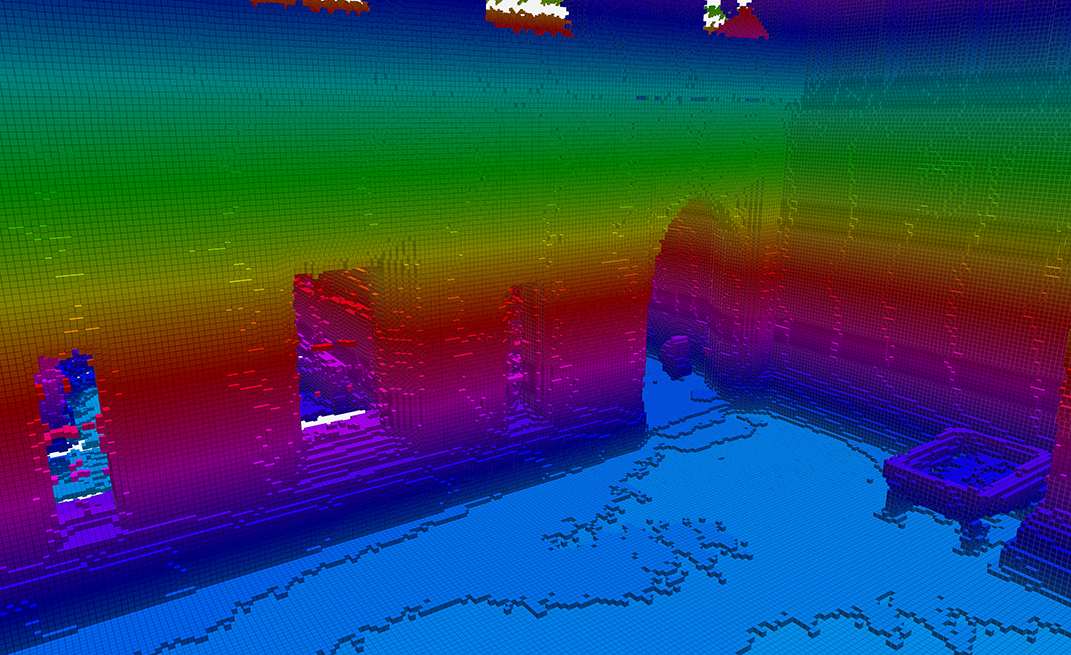} & 
 \includegraphics[width=0.25\textwidth]{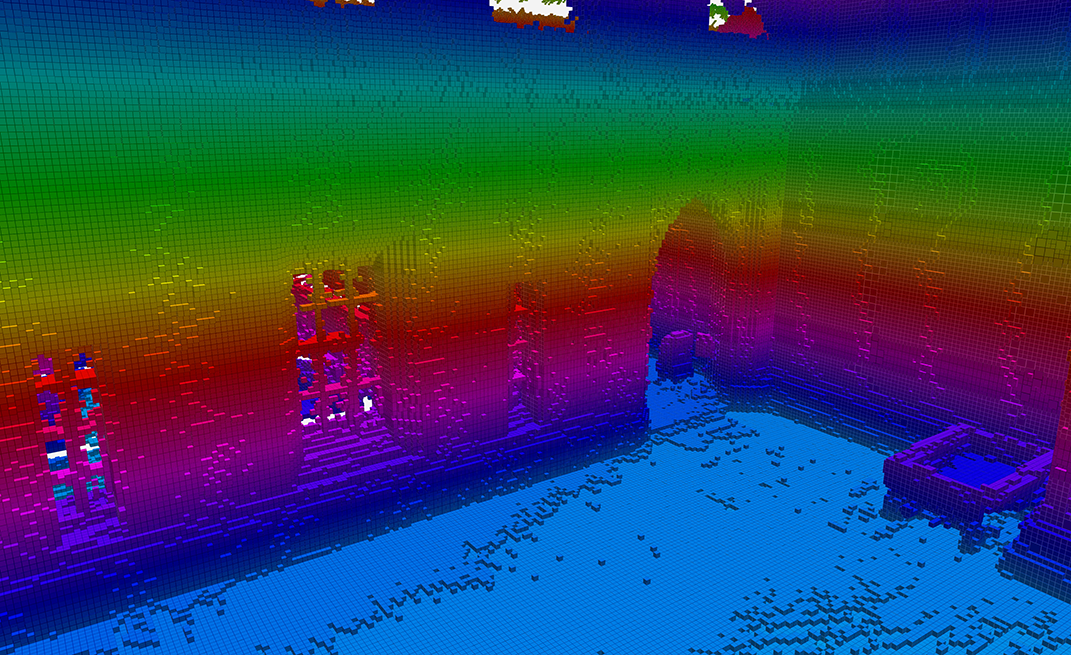}\\[3pt]
 octomap & supereight2 & voxblox & ours (beams)\\
\end{tabular}
\caption{Qualitative reconstruction comparisons featuring detailed geometry on scenes of the Panoptic mapping (top) and Newer College (bottom) datasets, both at 5cm resolution.}
    \label{fig:qualitative_evaluations}
    \vspace{-0.5em}
\end{figure*}

From the \ac{AUC} values shown in Table \ref{table:auc} we can see that, when using larger cell sizes, only our proposed method can compete with voxblox. Being TSDF-based, voxblox can more accurately reconstruct smooth surfaces which account for large parts of the environment, giving it a distinct advantage. The remaining two methods have worse overall performance. When moving to a higher resolution the difference shrinks and all methods perform comparably. Looking at the results shown in Figure \ref{fig:panoptic_flat_quantitative} we can clearly see where octomap and supereight2 accumulate their errors in the \SI{5}{\centi\meter} resolution case. Octomap struggles to properly localize the surface boundary, while supereight2 is overly pessimistic, labeling cells far from the surface as occupied. Finally, one can see the trade-off between our method, using a beam-based model, and voxblox, using a TSDF model. Voxblox has better at the surface reconstruction performance while our approach is better at reconstructing thin objects. This difference can also be seen in Figure \ref{fig:qualitative_evaluations} where the chair is missing its legs in the voxblox reconstruction. Looking at the \SI{2}{\centi\meter} resolution case, all methods but supereight2 perform almost identically. Supereight2 still produces pessimistic results, which likely stem from the approximations used to achieve its impressive speed.

\subsubsection{Newer College dataset}

The second set of experiments uses the Cloister sequence from Collection 2 in the Newer College dataset \cite{zhangMultiCameraLiDARInertial2022}. This sequence was chosen because it captures geometry with a wide range of scales including wide-open outdoor spaces, indoor spaces with arches and sculptures, and vegetation.
Odometry estimates and undistorted point clouds were obtained using FastLIO2 \cite{xuFASTLIO2FastDirect2022} processing the Ouster OS0-128 IMU and point cloud data. The motion-compensated point clouds were used for all frameworks except supereight2, which operates using dense range images and does not yet support motion-undistortion.

\begin{figure}[bt]
    \centering
    \includegraphics[width=0.475\textwidth]{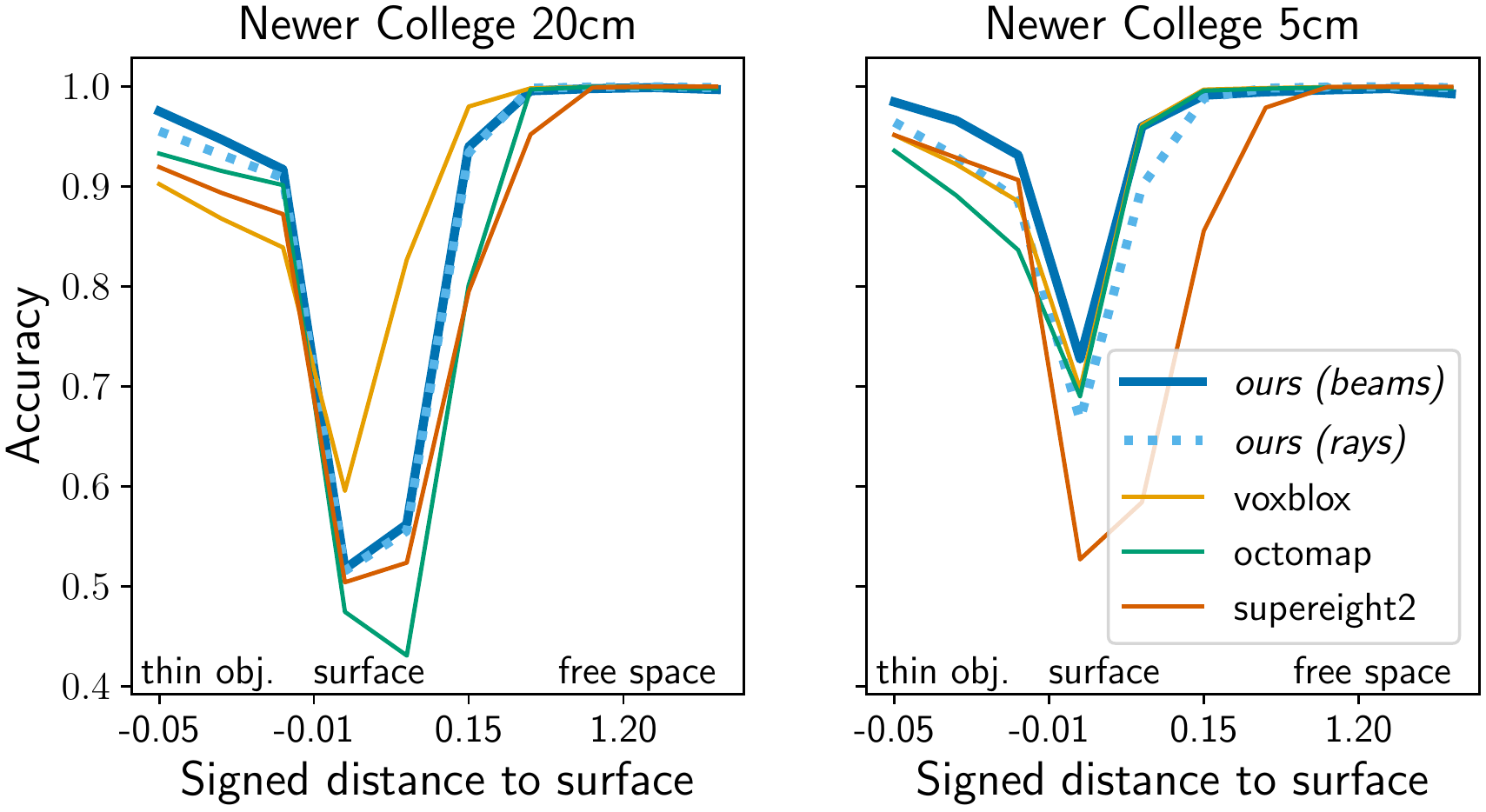}
    \caption{Accuracy in function of the distance to the surface on the Newer College dataset at different resolutions. Higher is better.}
    \label{fig:accuracy_over_distance_newer_college}
    \vspace{-1em}
\end{figure}

Looking at the \ac{AUC} numbers in Table \ref{table:auc} we immediately see that this real-world LiDAR dataset is more challenging than the previous synthetic one. When using a coarse \SI{20}{\centi\meter} resolution octomap performs the worst, with voxblox and our method achieving the best results, and supereight2 landing in the middle. Moving to a higher resolution of \SI{5}{\centi\meter} octomap and supereight2 end up performing similar while voxblox slightly outperforms our approach. However, the detailed results shown in Figure \ref{fig:accuracy_over_distance_newer_college} reveal interesting insights. At \SI{20}{\centi\meter} resolution octomap struggles to produce accurate surfaces. We also see that our approach and supereight2 have similar performance when it comes to reconstructing the surface but our approach performs slightly better when classifying free space in the vicinity of obstacles. Voxblox again performs the best in surface reconstruction and free space classification, but suffers in the thin object reconstruction domain. Moving to a finer \SI{5}{\centi\meter} resolution the change is similar to that observed in the Panoptic dataset. The accuracy of every method improves and they move closer together, with supereight2 failing to accurately predict free space close to surfaces. The differences between the other three methods are characterized by octomap not reconstructing thin objects accurately while both voxblox and ours (beams) perform equally well.

\subsubsection{Sensor model ablation}

To verify the benefit of the more costly uncertainty aware sensor model proposed in Section \ref{sec:measurement_model}, we conduct an ablation comparing our proposed sensor model, \emph{ours (beams)}, with one that disregards angular uncertainty, \emph{ours (rays)}. As the Panoptic Flat dataset contains no noise on observation or pose there is, as to be expected, no difference between the two models. On the Newer College dataset, however, there are visible differences. In the coarse setting the proposed uncertainty-aware model improves the ability to reconstruct thin objects. Moving to the higher resolution case both the ability to reconstruct surfaces and thin objects are significantly improved by our proposed model.

These accuracy evaluations showed several things. The proposed method \emph{ours (beams)} compares favorably to the other three methods. Despite the natural advantage voxblox has in surface reconstruction tasks, being a TSDF-based method, our approach performs on par while having superior performance in thin object reconstruction. The uncertainty-aware sensor model also improves the quality of the map close to surfaces and when dealing with thin surfaces, allowing the reconstruction of objects that other methods can't capture when using the same cell size.

\subsection{Efficiency evaluations}

We evaluate the memory usage as well as the runtime of our method in comparison to the three baseline methods. Memory usage is reported as the amount of RAM used by the method as well as the memory used by the map data structure. While our framework can be implemented using various data structures, we used octomap's octree implementation to keep the comparison as fair as possible. The runtime is reported as the elapsed wall time and the cumulative CPU time across all threads, allowing a fair comparison between single-threaded and multi-threaded methods. All frameworks have their visualizations disabled and all experiments are performed on the same desktop computer with an Intel i9-9900K CPU.

From the numbers shown in Table \ref{tab:comp_perf} we can see that supereight2 ranks first in terms of wall time on the depth camera dataset, and second best for LiDAR. However, the memory usage of its maps is relatively large owing to the fact that it estimates occupancy using weighted averaging instead of log-odds updates (requiring 2 floats per cell instead of 1) and focuses its implementation primarily on speed. Voxblox, as to be expected from a TSDF-based method, has the largest map sizes at higher resolutions but is computationally efficient. Octomap produces large maps, in comparison to our method, and is the slowest of all compared methods by an order of magnitude. Octomap's significant slowdown at high resolutions is caused by the fine-to-coarse model employed by their integrator which needs to touch every single cell. Our proposed method obtains maps that are significantly smaller than those of octomap despite using the same underlying data structure.

The runtime of our method, when looking at the CPU time, is equal or better than that of supereight2. However, as supereight2's implementation uses multiple threads the real-world performance of it is still better. The difference in runtime and memory usage between our method and octomap clearly shows the benefits of using wavelets to represent the map as it enables good compression and allows the use of an efficient coarse-to-fine integrator capable of skipping unnecessary work.

Comparing the memory and runtime of \emph{ours (rays)} and \emph{ours (beams)} we can see that the price for the improved quality is larger maps by about 30\% to 70\% depending on the resolution and an increase in runtime of around 50\%. These increases stem from the fact that the uncertainty-aware model needs to update more voxels and that the map contains more fine details and voxels with partial occupancy values.
Overall, our proposed method shows good general performance in both memory usage and runtime, with clear avenues for improvements. The wall time could be reduced significantly using multi-threading, which is easily achievable due to the independence of the voxel updates. Moreover, we believe the memory used to store the map itself could be reduced further by using a more efficient data structure such as the one proposed by OpenVDB \cite{musethOpenVDBOpensourceData2013}. These extensions will be added to the open-source code.

\begin{table}[bt]
    \caption{Computational resource usage at different resolutions. Lower is better.}
    \label{tab:comp_perf}
    \begin{tabular}{@{\,}l @{\quad}m{3mm}l@{\ \ }r@{\quad}r@{\quad}r@{\quad}r@{\,}}
        \hline
         &  &  & \multicolumn{2}{c}{Memory (MB)} & \multicolumn{2}{c}{Time (s)} \\
        Dataset & Res & Framework & RAM & Map only & CPU time & Wall time \\ \hline
        \multirow{8}{*}{\begin{tabular}[c]{@{}l@{}}Panop.\\Flat\end{tabular}} & \multirow{4}{*}{5cm} & octomap & 162.35 & 6.50 & 130.32 & 129.00 \\
         &  & supereight2 & 158.23 & 46.09 & 27.79 & \textbf{4.76} \\
         &  & voxblox & 229.96 & 36.90 & 58.58 & 10.68 \\
         &  & \textit{ours (rays)} & 135.69	& \textbf{4.17}	& \textbf{5.58}	& 6.78 \\
         &  & \textit{ours (beams)} & \textbf{130.04} & 5.65 & 6.94 & 7.20 \\ \cline{2-7} 
         & \multirow{4}{*}{2cm} & octomap & 6202.39 & 50.94 & 773.16 & 763.39 \\
         &  & supereight2 & 448.38 & 285.07 & 50.83 & \textbf{9.32} \\
         &  & voxblox & 663.53 & 348.15 & 244.69 & 24.61 \\
         &  & \textit{ours (rays)} & 343.26	& \textbf{39.09}	& \textbf{33.00}	& 34.80 \\
         &  & \textit{ours (beams)} & \textbf{294.58}	& 67.81	& 57.56	& 57.51 \\ \hline
        \multirow{8}{*}{\begin{tabular}[c]{@{}l@{}}Newer\\Coll.\end{tabular}} & \multirow{4}{*}{20cm} & octomap & 203.25 & 20.78 & 688.71 & 709.99 \\
         &  & supereight2 & 249.03 & 107.79 & 411.67 & 67.14 \\
         &  & voxblox & 261.02 & 66.32 & 228.12 & \textbf{48.07} \\
         &  & \textit{ours (rays)} & 180.86	& \textbf{6.94}	& \textbf{87.39}	& 88.78 \\
         &  & \textit{ours (beams)} & \textbf{138.92} & 8.82 & 107.67 & 113.26 \\ \cline{2-7} 
         & \multirow{4}{*}{5cm} & octomap & 14404.76 & 981.02 & 36252.70 & 35790.60 \\
         &  & supereight2 & 2926.42 & 2333.93 & 2853.12 & 404.19 \\
         &  & voxblox & 3718.85 & 2362.58 & 1788.90 & \textbf{162.36} \\
         &  & \textit{ours (rays)} & 1192.95	& \textbf{241.84}	& \textbf{1656.26}	& 1671.58 \\
         &  & \textit{ours (beams)} & \textbf{1065.21} & 402.18 & 2085.05 & 2083.61 \\ \hline
    \end{tabular}
    \vspace{-1.5em}
\end{table}

\subsection{Multi-sensor multi-resolution mapping}

\begin{figure}
    \centering
    \subfloat{\includegraphics[width=0.16\textwidth]{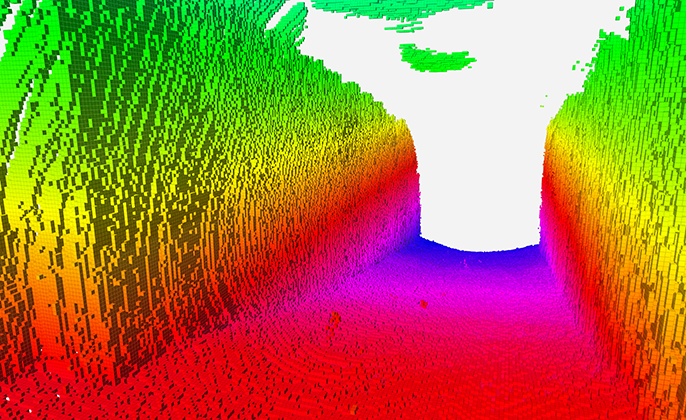}}
    \hfill
    \subfloat{\includegraphics[width=0.16\textwidth]{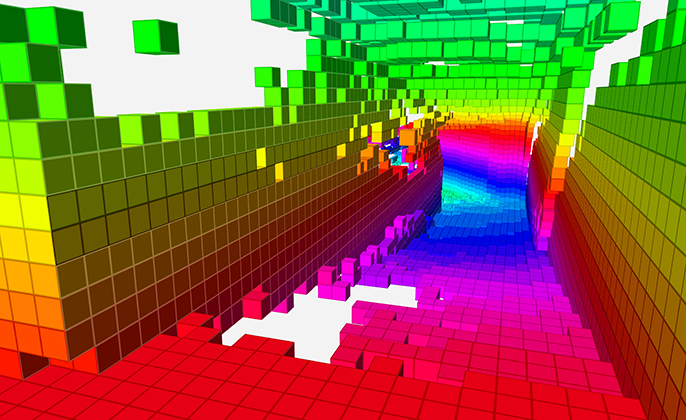}}
    \hfill
    \subfloat{\includegraphics[width=0.16\textwidth]{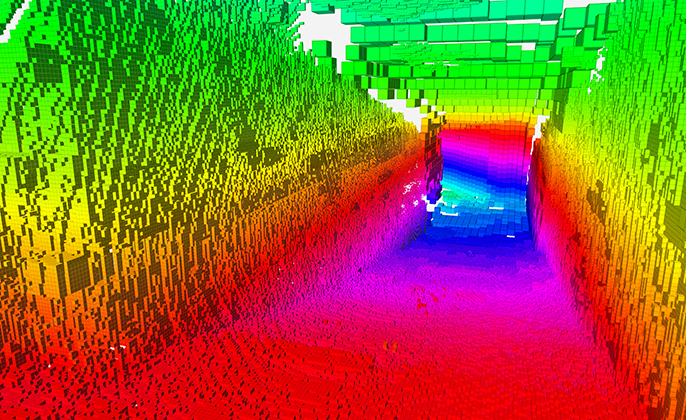}}
    \caption{Example of our framework performing multi-sensor, multi-resolution volumetric mapping on the DARPA SubT Finals dataset, combining data from 2 ground-facing LiDARs at 2cm resolution (left) and 1 horizontal LiDAR at 16cm resolution up to a range of 30m (center) into a single map (right).}
    \label{fig:darpa_subt_multi_sensor_multi_res}
    \vspace{-1em}
\end{figure}

One key advantage of our framework is its natural ability to handle multiple sensors at different resolutions. In this experiment, we show qualitative results of our mapping framework running in multi-sensor mode on the DARPA SubT Finals dataset \cite{tranzattoTeamCERBERUSWins2022}. In Figure \ref{fig:darpa_subt_multi_sensor_multi_res} we show the output of our framework simultaneously integrating two Robosense Bpearl dome-LiDAR sensors and one Velodyne VLP-16 LiDAR. The Bpearls were angled to scan the ground around the robot while the VLP-16 was providing long-range observations. 
As the sensors provide information for different purposes we integrate them with different resolutions into the map. The Bpearls, responsible for local terrain mapping to enable navigation of a quadruped, are integrated at \SI{2}{\centi\meter} resolution. At the same time the VLP-16, responsible for long-range mapping and exploration, is integrated with a resolution of at most \SI{16}{\centi\meter}. This results in a unified map, that supports both local trajectory planning as well as global exploration goal placements, without wasting resources on high-accuracy map reconstruction in areas where it is not needed. While shown here for multiple LiDAR sensors the same approach can be used for mobile manipulation setups using a 3D LiDAR for navigation and RGB-D cameras for scene reconstruction, resulting in a map that supports both navigation as well as manipulation.

\section{Conclusion}

In this work, we introduced \textit{wavemap}, a hierarchical volumetric mapping framework inspired by multi-resolution analysis. The \ac{MRA} theory guarantees that using wavelet decomposition, we can safely and very efficiently integrate new observations in a coarse-to-fine manner. The resulting gains in computational efficiency, together with early stopping criteria for the integrator, allow us to use more complex sensor models such as the proposed angular uncertainty-aware model. In experiments on synthetic RGB-D and real-world 3D LiDAR data, we demonstrate that our proposed method achieves high-quality results while being efficient in terms of memory and compute requirements. We also demonstrate how our method can incorporate observations from multiple sensors into a single map with per-sensor resolution. This allows the use of a single map representation for tasks that would have required several dedicated maps in the past. Finally, we open source the implementation of our approach to facilitate future research.

\section*{Acknowledgments}
This work has received funding from the European Union’s Horizon 2020 research and innovation programme under grant agreement No 871542.

\bibliographystyle{plainnat}
\bibliography{bibliography/references}

\appendix
\section{A brief introduction to wavelets}
This appendix is intended as a primer on wavelet theory, providing additional context for the method section. We start with a short introduction to the \ac{MRA} conditions, before showing how orthogonal wavelet bases fulfill these requirements. We then discuss how wavelet decompositions can efficiently be computed using the Fast Wavelet Transform.
For readers that are interested in learning more about sparse signal processing using wavelets, we warmly recommend \cite{daubechiesTenLecturesWavelets1992,mallatWaveletTourSignal2009}.

\subsection{Multi-Resolution Analysis}
\label{appendix:mra}
% Consider using the MRA explanation by Daubechies p14 that skips Riesz basis and directly requires V_0 admits a basis consisting of shifted scaling functions.
% See also Daubechies p129
Multi-resolution representations are regularly used in the context of computer vision and robotics. For example, in Laplacian image pyramids introduced by Burt and Adelson \cite{burtLaplacianPyramidCompact1983}. Mallat and Meyer \cite{mallatWaveletTourSignal2009}, formalized the expected behavior of multi-resolution representations through the \ac{MRA} conditions:
\begin{align}
    \forall(j,k)\in\mathbb{Z}^2,&&  f(x) \in V_j &\Leftrightarrow f(x - 2^j k) \in V_j      \label{eq:mra_1} \\
    \forall j \in\mathbb{Z},&&      V_{j+1}      &\subset V_j                               \label{eq:mra_2} \\
    \forall j \in\mathbb{Z},&&      f(x) \in V_j &\Leftrightarrow f(x/2) \in V_{j+1}        \label{eq:mra_3} \\
    \span\span \lim_{j\to\infty}V_j   = \bigcap_{j=-\infty}^{\infty} V_j = \{0\} \span      \label{eq:mra_4} \\
    \span\span \lim_{j\to -\infty}V_j = \text{closure}\left(\bigcup_{j=-\infty}^{\infty}V_j\right) = L^2(\mathbb{R}) \span \label{eq:mra_5} \\
    \span\span\text{$V_0$ admits a Riesz basis}\span \label{eq:mra_6}
\end{align}
where the sequence of subspaces $\{V_j\}_{j\in\mathbb{Z}}$ corresponds to the map's representations at increasing resolution levels $2^j$, referred to as scales, and  each $V_j$ is a closed subspace of Lebesgue space $L^2$. Starting with condition \ref{eq:mra_6}, the most common Riesz basis used in robotics consists of box functions arranged to span the cells of a regular grid. In this case, the scale $2^j$ corresponds to the cell width.
Condition \ref{eq:mra_1} ensures self-similarity in space. Specifically, if subspace $V_j$ can represent function $f(x)$, it can also represent the same function shifted by integer multiple of the cell size.
Condition \ref{eq:mra_2} states that the subspaces are nested. In other words, any function contained in subspace $V_{j+1}$ must also be contained in next finer subspace $V_j$ and by extension in all finer subspaces.
Condition \ref{eq:mra_3} ensures self-similarity in scale. If $V_j$ contains $f(x)$, $V_{j+1}$ must be able to contain $f(x)$ dilated by $2$.
Finally, conditions \ref{eq:mra_4} and \ref{eq:mra_5} ensure completeness. At the coarsest scale ($j\to\infty$), $V_j$ only contains the zero element, whereas refining the scale ($j\to-\infty$) eventually allows us to represent any signal in $L^2$.

\subsection{Orthogonal wavelet bases}
\label{appendix:orthogonal_wavelet_bases}
The principal idea behind wavelets is that they represent the difference between the consecutive resolutions of a signal's \ac{MRA}.
Formally, they span a second subspace $W_j$ which is the orthogonal complement to $V_j$, such that $V_j \oplus W_j = V_{j-1}$ where $\oplus$ is the vector-space direct sum operator. In words, this means that by combining a signal's representation $V_j$ with its wavelet details at the same resolution $W_j$ we obtain the signal's representation at the next higher resolution $V_{j-1}$.

An orthogonal basis for all $V_j$ can be obtained by translating and dilating a single function $\phi$, referred to as the scaling function, as $\phi_{jk}(x)=\frac{1}{2^j}\phi(\frac{x-2^jk}{2^j})$. The scaling function can be found by orthogonalizing the Riesz basis of $V_0$ as described in \cite{mallatWaveletTourSignal2009}. In similar fashion, an orthogonal basis for $W_j$ can be obtained by translating and scaling a single wavelet function $\psi$ as $\psi_{jk}(x)=\frac{1}{2^j}\psi(\frac{x-2^jk}{2^j})$. One condition that any wavelet function has to fulfill in order to be admissible is that its average must be zero $\int_{-\infty}^{\infty} \psi(x) dx = 0$. More generally, the scaling functions and wavelet functions can be seen as complementary low and high-pass filters that, when combined, can perfectly reconstruct the signal from the next finer scale.
Since wavelet bases form a valid \ac{MRA}, this concept can be applied recursively and the entire map can be represented by stacking a single scaling function at the coarsest scale with a hierarchy of wavelet functions at increasing scales.

Note that the Riesz basis consisting of box filters arranged to span the cells of a regular grid, mentioned previously, is already orthogonal. In fact, the unit box filter can be used as a scaling function
\begin{equation}
    \phi(x) =
    \begin{cases} 
    1 & 0 \leq x < 1 \\
    0 & \text{otherwise}
    \end{cases}    
\end{equation}
and doing so directly leads to the Haar basis \cite{mallatWaveletTourSignal2009}. The corresponding Haar wavelet function can be derived by finding $\phi$'s orthogonal complement while enforcing the \ac{MRA} conditions and is given by
\begin{equation}
    \psi(x) =
    \begin{cases} 
    -1 & 0 \leq x < 1/2 \\
    1 & 1/2 \leq x < 1 \\
    0 & \text{otherwise}
    \end{cases}
\end{equation}
Orthogonal wavelet bases of $\mathbb{R}$ can be extended to separable orthogonal bases $b$ for $\mathbb{R}^n$ by combining the scaling and wavelet functions along each dimension as 
\begin{equation}
b = \left\{\prod_{k=1}^n \phi(x_k)^{o_k} \psi(x_k)^{1-o_k} \right\}_{\forall o \in \{0,1\}^n}    
\end{equation}

% [Figure showing the Haar basis functions in 2D.]

\subsection{The Fast Wavelet Transform}
\label{appendix:fast_wavelet_transform}
The discrete wavelet transform for a function $f$ and wavelet $\psi$ is defined as the projection of $f$ onto the set of all integer scalings and translations of the wavelet function $\{\psi_{jk}\}_{j,k\in\mathbb{Z}}$.
Each wavelet coefficient $d_{jk}$ is thus computed as
\begin{equation}
    d_{jk} = \sum_{x=-\infty}^{\infty}f(x) \frac{1}{2^j}\psi\left(\frac{x - 2^j k}{2^j}\right)
\end{equation}
where the summation could be replaced by an integral if the domain of $f$ is real-valued instead of discrete. Note that this transform is linear and, for orthogonal wavelets, orthogonal.

The coefficients $d_{jk}$ can efficiently be computed using the \ac{FWT} algorithm \cite{mallatWaveletTourSignal2009}, which exploits the hierarchical \ac{MRA} structure to remove redundant operations. The \ac{FWT} is initialized by projecting $f$ onto the scaling functions at the finest scale $a_{0k} =  \sum_{-\infty}^{\infty}f(x) \phi\left(x -  k\right)$ or with a good approximation thereof. At each iteration, these coefficients are then filtered and downsampled to obtain the wavelet and scaling coefficients at the next coarser scale. These iterations are typically repeated until only $1$ scaling coefficient is left or a desired number of levels is reached. For wavelets with finite spatial support and functions $f$ sampled at $N$ points, the \ac{FWT} computes the full wavelet decomposition in $O(N)$ time.
Extending the \ac{FWT} to only (de)compress regions-of-interest or single cells is straightforward and very efficient if the spatial support of the chosen wavelet is small, as is the case for the Haar wavelet.

\end{document}